\documentclass[runningheads]{llncs}
\usepackage{graphicx}

\usepackage{booktabs}
\usepackage{amsmath}

\usepackage{cleveref}
\usepackage[caption=false]{subfig} 

\begin{document}
\title{The Power of Training: How Different Neural Network Setups Influence the Energy Demand}
\titlerunning{The Power of Training}
%
\author{Daniel Geißler\inst{1}
\and
Bo Zhou\inst{1,2}
\and
Mengxi Liu\inst{1}
\and
Sungho Suh\inst{1}
\and 
Paul Lukowicz\inst{1,2}
}
\authorrunning{D. Geißler et al.}
%
\institute{German Research Center for Artificial Intelligence (DFKI),  Kaiserslautern, Germany \and
University of Kaiserslautern-Landau (RPTU), Kaiserslautern, Germany}

\maketitle              

\begin{abstract}

This work offers a heuristic evaluation of the effects of variations in machine learning training regimes and learning paradigms on the energy consumption of computing, especially HPC hardware with a life-cycle aware perspective.
While increasing data availability and innovation in high-performance hardware fuels the training of sophisticated models, it also fosters the fading perception of energy consumption and carbon emission.
Therefore, the goal of this work is to raise awareness about the energy impact of general training parameters and processes, from learning rate over batch size to knowledge transfer. 
Multiple setups with different hyperparameter configurations are evaluated on three different hardware systems.
Among many results, we have found out that even with the same model and hardware to reach the same accuracy, improperly set training hyperparameters consume up to 5 times the energy of the optimal setup.
We also extensively examined the energy-saving benefits of learning paradigms including recycling knowledge through pretraining and sharing knowledge through multitask training.
\end{abstract}
\section{Introduction}
With the wake of artificial intelligence (AI), the high-performance computing (HPC) demand related to machine learning has been drastically elevated, together with the energy consumption in the sector as a whole.
While there have been many works optimizing the energy footprint of machine learning from the software model architecture and hardware accelerator aspects, there has been little effort in examining the differences in training regimes and learning paradigms, especially considering the lifecycle of machine learning models.
The connection between the complexity of a task being solved by a deep neural network (DNN) and the accompanying energy requirement from powerful hardware to train and deploy such models is a well-known issue of machine learning algorithms.
Improving availability and access to more data, which forms the basis for training complex models, supports the trend towards enhanced hardware power and energy consumption further.
Next to the increase in computational demands, there is a growing awareness of a less visible but increasingly significant aspect: the carbon footprint for training machine learning models. \cite{henderson2020towards,patterson2022carbon}

Multiple projects like carbontracker \cite{anthony2020carbontracker} or eco2AI \cite{budennyy2022eco2ai} apply to this problem statement, aiming to track the energy consumption and the related carbon footprint for training a model by linking into the hardware power utilization logging during the model training.
Although these tools can monitor energy consumption, they are unable to compare the efficiency of the training process as well as suggesting recommendations for improvements.

In this work, we aim to create awareness about the significance of general adjustments in the training process and their major effects on energy consumption.
Moreover, a goal of this work is to strive for a general analysis that can be applied to other projects as well.
Additionally, we examine the effects of advanced learning paradigms to project our baseline findings from standard training regimes into state-of-the-art practices by comparing their energy consumption on our test setups.

Even though the innovation in terms of efficiency and sustainability of machine learning algorithms lags behind the quantitative growth of hardware resources, small improvements in the low percentage range already help to reduce the carbon footprint if such methods are scaled properly within the landscape of machine learning applications. \cite{patterson2022carbon}

In short summary, we make the following contributions:

\begin{enumerate}
    \item We examine the energy consumption of often overlooked nuance during DNN training, specifically among training setup hyperparameters and learning paradigms with longer lifecycle considerations. A major factor differentiating our method is that we use the prediction accuracy as a performance target, resulting in total energy per development cycle, additionally to the pure energy per epoch metrics like most related works.
    \item Our findings show, with the same DNN model, HPC hardware, and accuracy-regulated tasks, improperly configured batch size and learning rate could cause 5 times the energy consumption of the optimal configurations.
    \item By re-using part of a DNN (the feature encoder) in different downstream classification tasks, the practice will bring a break-even point compared to training a new model in total for every downstream task. And this break-even point is associated with the training hyperparameter setups in Contribution No. 2.
    \item By sharing parts of a DNN during the same training cycle of multiple tasks (multi-task training), the energy consumption per cycle is significantly less than the total of training for different tasks in isolation.
\end{enumerate}

The paper is structured as follows: \cref{sec:sota} reviews the relevant state-of-the-art works to track energy consumption and optimize the hyperparameter space; \cref{sec:method} introduces the methodology of conducted experiments from classic training regime to advanced learning paradigms followed by the corresponding results for the classic training in \cref{sec:regime_results}, the pretraining scenario in \cref{sec:pre_results} and multitask scenario in \cref{sec:mutli_results}.
After additional comparison of the HPC results to high-performance consumer hardware in \cref{sec:prosumer}, we conclude with a discussion on limitations and future work in \cref{sec:conlusion}.

\section{Related Works}
\label{sec:sota}
Due to lacking practicability and the complexity of tasks, precise energy consumption tracking is still not commonly applied in the field of AI and Machine Learning \cite{garcia2019estimation}.
However, the rising demand for energy must be counteracted to minimize the environmental footprint and economic expenses, for instance when hosting large server and HPC farms, in order to operate sustainably in the future \cite{narciso2020application}.
In the past focus lay on inference optimization for edge devices, microcontrollers, and field-programmable gate arrays (FPGAs), since the energy demands of training machine learning models have received less attention due to smaller implications \cite{daghero2021energy}.
Nevertheless, with the rise of Large Language Models (LLM) and other complex architectures, the computational demands for training such models have a significantly larger share of the lifecycle energy requirement \cite{strubell2019energy}.
This in turn requires a precise consideration of possible adjustment screws to counteract these trends, whereas this work focuses on the exploration and tuning of hyperparameters \cite{he2019control}.
To the best of our knowledge, this work is the first to investigate the selection of popular hyperparameter spaces with a hardware-based focus on quantifying the energy demand instead of purely aiming towards optimal model performance.

\subsection{Energy Tracking}
Energy tracking software tools are usually designed to capture as much power information about the system as possible by building an additional layer between the system's hardware configuration and the user's model training process.
Generally speaking, the power consumption of the Graphics Processing Unit (GPU) is the largest part of the training process as it performs the core work with the parallel processing of mathematical tasks.
Following that, usually, the Central Processing Unit (CPU) is the second largest consumer of power.

There is a growing list of software available with most of them utilizing the same approach to gather the power consumption data from the hardware manufacturers' utility logging.
The energy consumption is then calculated from the current power consumption and the polling time interval set in the software.
Works like Carbontracker \cite{anthony2020carbontracker}, eco2Ai \cite{budennyy2022eco2ai} and Green Algorithms \cite{lannelongue2021green} utilize this approach to gather data from CPU, GPU, and even the memory if supported.
Nevertheless, there is still a lack of tracking the full systems' energy consumption.
High-performance setups may consume more power than present tools can track, for instance, if the cooling, which can account for a non-negligible proportion, is not taken into account.
Therefore, software like Carbontracker \cite{anthony2020carbontracker} multiplies its results with an efficiency constant of 1.55 to incorporate untracked secondary power needs and efficiency losses.

With an extended focus on user experience, projects like Cloud Carbon \cite{cloudCarbon} or CodeCarbon \cite{CodeCarbon} extend the gathered knowledge and present it in analytic-based dashboards.
Based on the calculated energy consumption and the user's location, the average local energy mix from fossil and renewable energy sources is utilized to estimate the carbon emissions in kilograms or even tons \cite{lacoste2019quantifying}.
On top of that, since the carbon emissions are difficult to visualize or imagine, the conversion into kilometers driven by car, flights with a plane, or the number of phones charged is a standard practice to make the user aware of the generated carbon emissions amount.

\subsection{Hyperparameter Selection}
Selecting the right hyperparameters is a common problem in thoroughly training the machine learning model to properly solve the targeted task.
It is still a standard practice to explore the hyperparameter space manually by launching multiple training runs and wasting energy on training insufficient models \cite{yang2020hyperparameter}.

Many techniques evolved over time to solve this issue, starting from classical methods like grid search or random search, towards more advanced techniques like evolutionary, Bayesian, pruning or reinforcement-learning based optimization strategies \cite{bischl2023hyperparameter}.
In many cases, there is no clear identification of the best-performing strategy possible because the efficiency of the algorithm depends on the respective machine learning problem to solve and the user's preferences.
Nevertheless, there is a trend of different hyperparameter optimization categories being fused to elevate the performance compared to the traditional algorithms \cite{bischl2107hyperparameter}.

All strategies together share a downside: They purely focus on maximizing the model performance without incorporating the energy consumption \cite{frey2022energy}. 
This oversight leads to an enhanced waste of energy and usually leads to training and discarding models that do not meet the required performance \cite{yu2020hyper}.
Therefore, there is a need to investigate the correlation between hyperparameter selection and energy demand in order to develop strategies that may strive for the best performance while maintaining moderate energy demands.

\section{Methodology}
\label{sec:method}
\subsection{Experiment Setup}
In order to provide a meaningful correlation between the training setup of a model and its power consumption, we chose two application scenarios that determine the structure of this work: computer vision and sensor-based activity recognition.
We used two commonly used benchmark datasets from two disciplines: for a computer vision task we used the CelebA dataset \cite{liu2015faceattributes} and for the sensor-based activity recognition, we utilized the PAMAP2 dataset \cite{reiss2012introducing} as a basis.

To reduce hardware-specific implications from our results and cross-compare them in a meaningful way, we run each experiment two times on three different hardware configurations on a SLURM-based HPC cluster \cite{yoo2003slurm}.
Due to the job-based nature of the HPC, the experiments run in clean, isolated Linux environments running on CUDA 12.3.
We settled three different Nvidia GPUs, namely the H100-HBM3 with 80GB, the A100-SXM4 with 80GB, and the GeForce RTX 3090.

The already introduced Carbontracker \cite{anthony2020carbontracker} is utilized to track the GPU power and energy consumption of the desired GPUs.
Due to Carbontracker's restriction on tracking CPU power consumption for Intel CPUs only and a lack of an adequate framework to track the power consumption of AMD CPUs, this work solely focuses on GPU power consumption.
The tracker is modified to additionally store the corresponding timestamp and the current epoch to properly link the gathered data with the model's training process.
The efficiency constant multiplication was disabled throughout our tests to achieve the raw GPU measurements.

To further classify the HPC results on common prosumer hardware, the same experimental scenarios were carried out again on a workstation equipped with an Nvidia RTX 6000 Ada with 48GB VRAM and a 16-inch Apple Macbook Pro with M1 Max chip (32-core GPU and 64GB unified memory).
The findings for those hardware are separately discussed in \Cref{sec:prosumer}.

Since each hardware setup has deviating power capabilities and the power is not representative in terms of energy consumption, we decided to calculate the epoch-wise energy consumption as the baseline following the equation:

\begin{equation}
\forall \, k \in [0, T] \quad E_{k} =  \frac{\sum_{i=0}^{n} power(k,n)}{n} * \frac{time(k)}{3600}
\label{equation1}
\end{equation}
\vspace{1px}

Throughout all experiments and the varying hardware setups, we kept the same model and training script versions.
Moreover, the changed parameters across training runs were induced as external arguments.
For training, we used early stopping techniques based on validation loss for the encoder and the validation accuracy for the classifier as a standard metric.

\subsection{Training Regime}
Our test procedure can be separated into two tasks which we applied to the two application scenarios, the training regime and the learning paradigm.
A common problem when training a model is the proper initialization of hyperparameters that fit the task or problem.
The two most prominent ones are the batch size and the learning rate \cite{he2019control}.
To determine the influence of hyperparameters, we created an encoder plus classifier architecture for the CelebA to classify the attributes and identities. 
The encoder is ResNet18 from the TIMM python package with pre-trained ImageNet weights and the classifier is based on classic convolution with a size of 20520 parameters.
For the PAMAP2, we build a similar convolutional classifier architecture inspired by \cite{suh2023tasked} with a size of 141293 parameters to classify the activities.
For the training regime, we trained the models with variations in batch size and learning rate, resulting in a total of 16 different setups as shown in \Cref{tab:parameters}.
We kept the hyperparameters constant for each training run without any further decay or adaption.
All experiments were conducted with threshold early stopping as a regularization in order to still maintain proper model performance.
We set the threshold to an accuracy of 0.8 from which the regularization was started.
It was set based on experience since such classifiers trained on CelebA and PAMAP2 usually range slightly above this performance.
This strategy ensures that all setups can exceed the threshold and further eliminates potential outliers due to early termination with poor accuracy.

\begin{table}[ht]
 \caption{Training Regime Hyperparameter Setup List}
  \label{tab:parameters}
  \centering
    \vspace{5px}
  \begin{tabular}{ccc}
    \toprule
    \textbf{Setup} & \textbf{Batch Size} & \textbf{Learning Rate} \\
    \midrule
    1 & 64    & 0.1 \\
    2 & 64    & 0.01 \\
    3 & 64    & 0.001 \\
    4 & 64    & 0.0001 \\
    5 & 256   & 0.1 \\
    6 & 256   & 0.01 \\
    7 & 256   & 0.001 \\
    8 & 256   & 0.0001 \\
    9 & 1024  & 0.1 \\
    10 & 1024 & 0.01 \\
    11 & 1024 & 0.001 \\
    12 & 1024 & 0.0001 \\
    13 & 4096 & 0.1 \\
    14 & 4096 & 0.01 \\
    15 & 4096 & 0.001 \\
    16 & 4096 & 0.0001 \\
    \bottomrule
  \end{tabular}
 
\end{table}

\subsection{Learning Paradigm}
On top of the training regime tests, we extended the experiments to explore the energy consumption of different learning paradigms focusing on knowledge sharing and reusing.
In principle, these learning paradigms can potentially reduce the energy cost compared to the baseline of training a randomly initialized, black-box model for each dataset or task.
Instead, the research community focuses on the generalization abilities or accuracy improvements of these learning paradigms; the actual energy consumption has not yet been evaluated.

\begin{figure}
    \centering
    \includegraphics[width=0.8\linewidth]{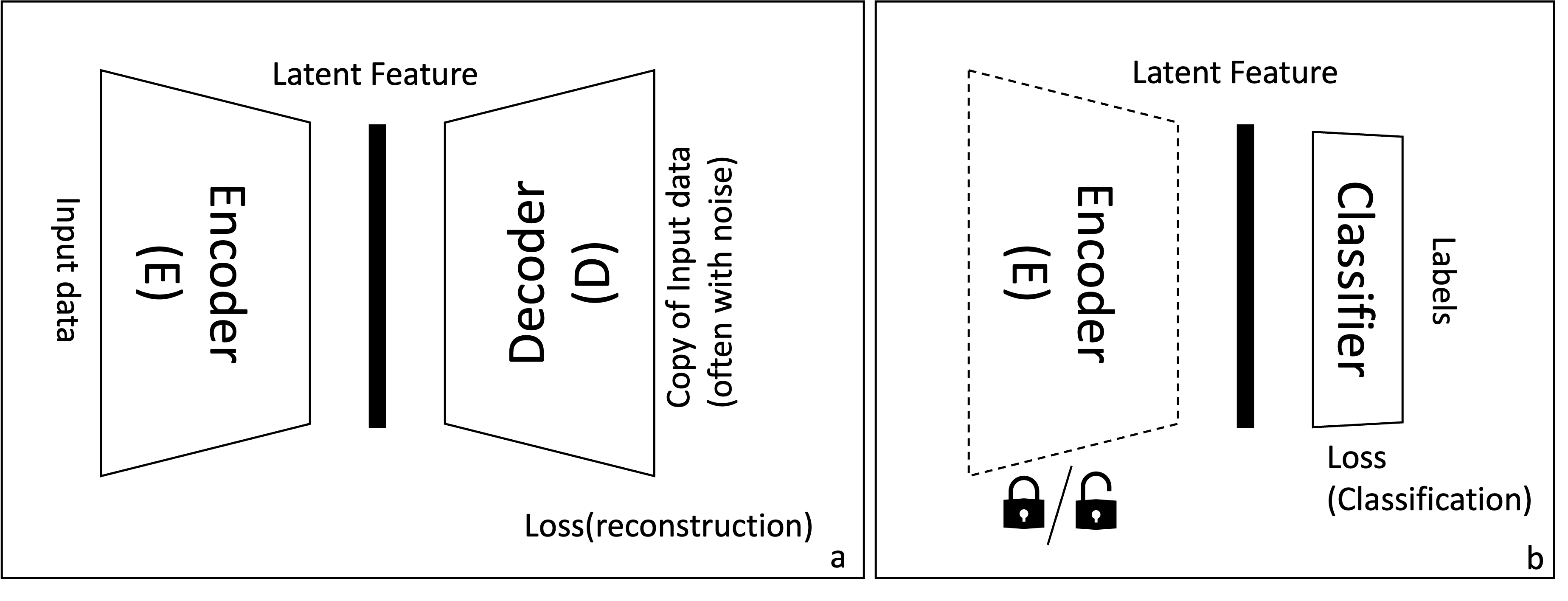}
    \caption{Pretraining an autoencoder and utilizing the encoder in frozen and unfrozen mode to generate latent features for the training of a classifier.}
    \label{fig:freeze}
\end{figure}

With the PAMAP2 dataset, we evaluate the pretraining learning paradigm, shown in \Cref{fig:freeze}. 
We train an autoencoder and use its encoder as a pre-trained part to train a classifier on its latent features and compare it to the training of the encoder and classifier architecture sequentially.
This practice is typically used to first force the feature encoder to learn information that can be used to reconstruct the input, thus the features are unique to different characteristics of the input data.
This could reduce the likelihood of over-fitting on the ground truth labels of the training data and thus provide a better-generalized model.
We also inspect the differences between freezing and unfreezing the pre-trained encoder weights during the usage for generating the latent features.

\begin{figure}[ht]
    \centering
    \includegraphics[width=0.8\linewidth]{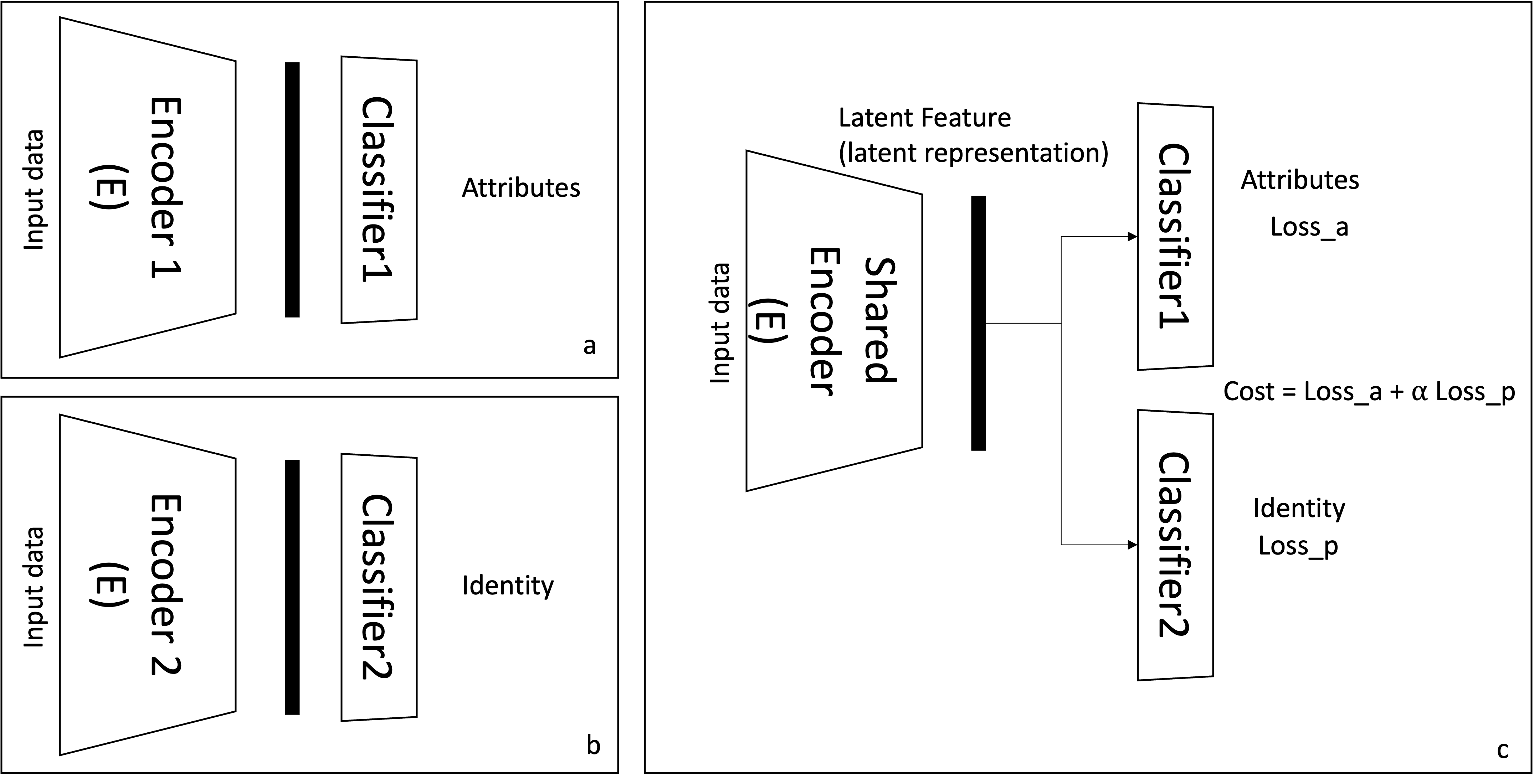}
    \caption{Comparing the single training setup with the multitask setup (combined loss function).}
    \label{fig:multitask}
\end{figure}

With the CelebA dataset, we can evaluate multi-task learning \cite{yin2017multi}, since the datasets have different aspects of ground truth for the same input (identity and facial attributes).
Opposed to training a complete black-box-like neural network for each task, multi-task learning shares the feature encoder with multiple downstream tasks (e.g. different classifiers).
The procedure is visualized in \Cref{fig:multitask} a and b for the single training setup and the multitask setup in c by connecting the two classifiers via the weight coefficient $\alpha$ within the loss function.

\section{Training Regime Results}
\label{sec:regime_results}

\begin{figure}[!ht]
\begin{minipage}[t]{0.32\textwidth}
  \centering
  \subfloat[CelebA on H100]{\includegraphics[width=\linewidth]{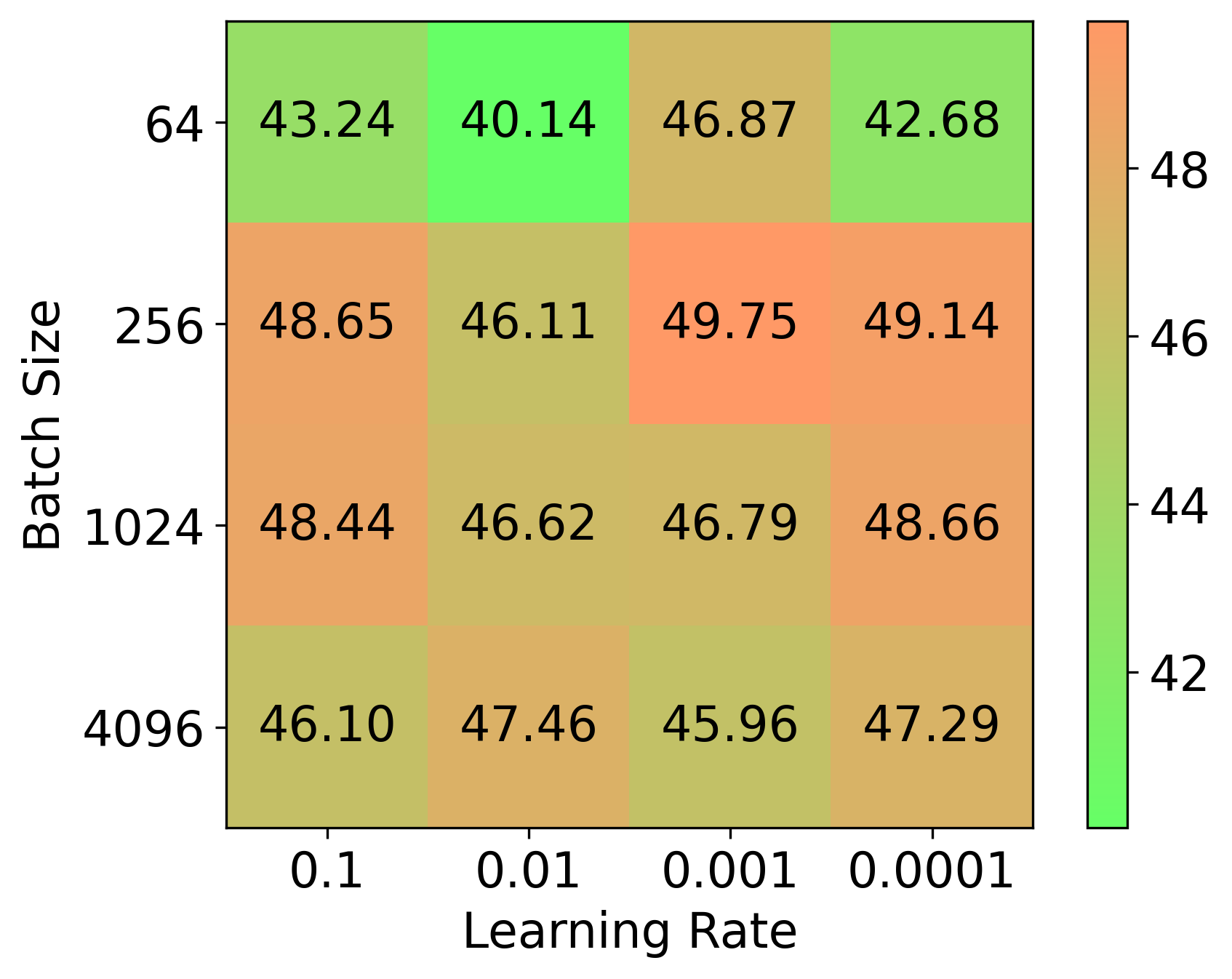}\label{fig:cel_H100_1}}
\end{minipage}
\hfill
\begin{minipage}[t]{0.32\textwidth}
  \centering
  \subfloat[CelebA on A100]{\includegraphics[width=\linewidth]{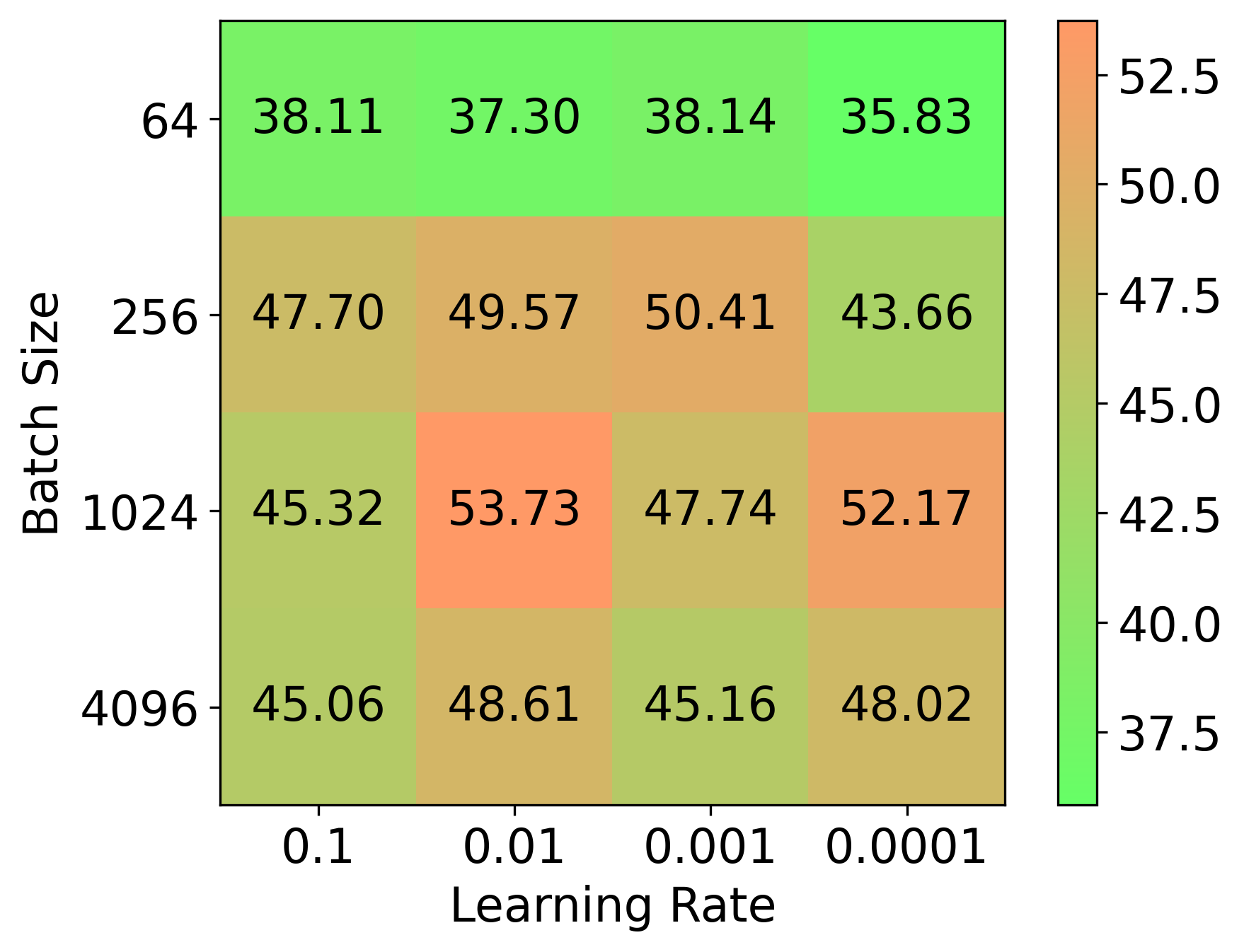}\label{fig:cel_A100_1}}
\end{minipage}
\hfill
\begin{minipage}[t]{0.32\textwidth}
  \centering
  \subfloat[CelebA on RTX 3090]{\includegraphics[width=\linewidth]{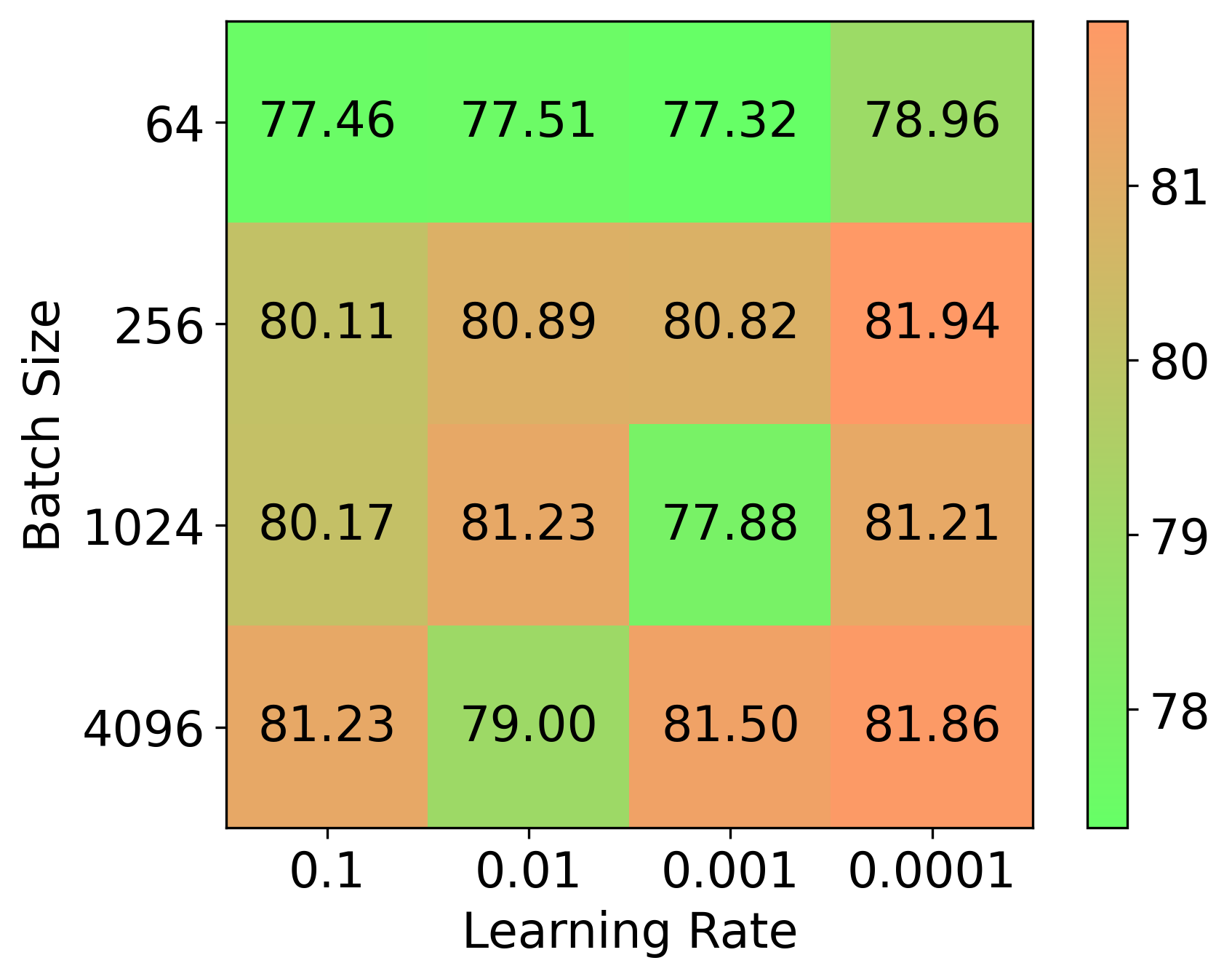}\label{fig:cel_rtx_1}}
\end{minipage}

\vspace{5px}

\begin{minipage}[t]{0.32\textwidth}
  \centering
  \subfloat[PAMAP2 on H100]{\includegraphics[width=\linewidth]{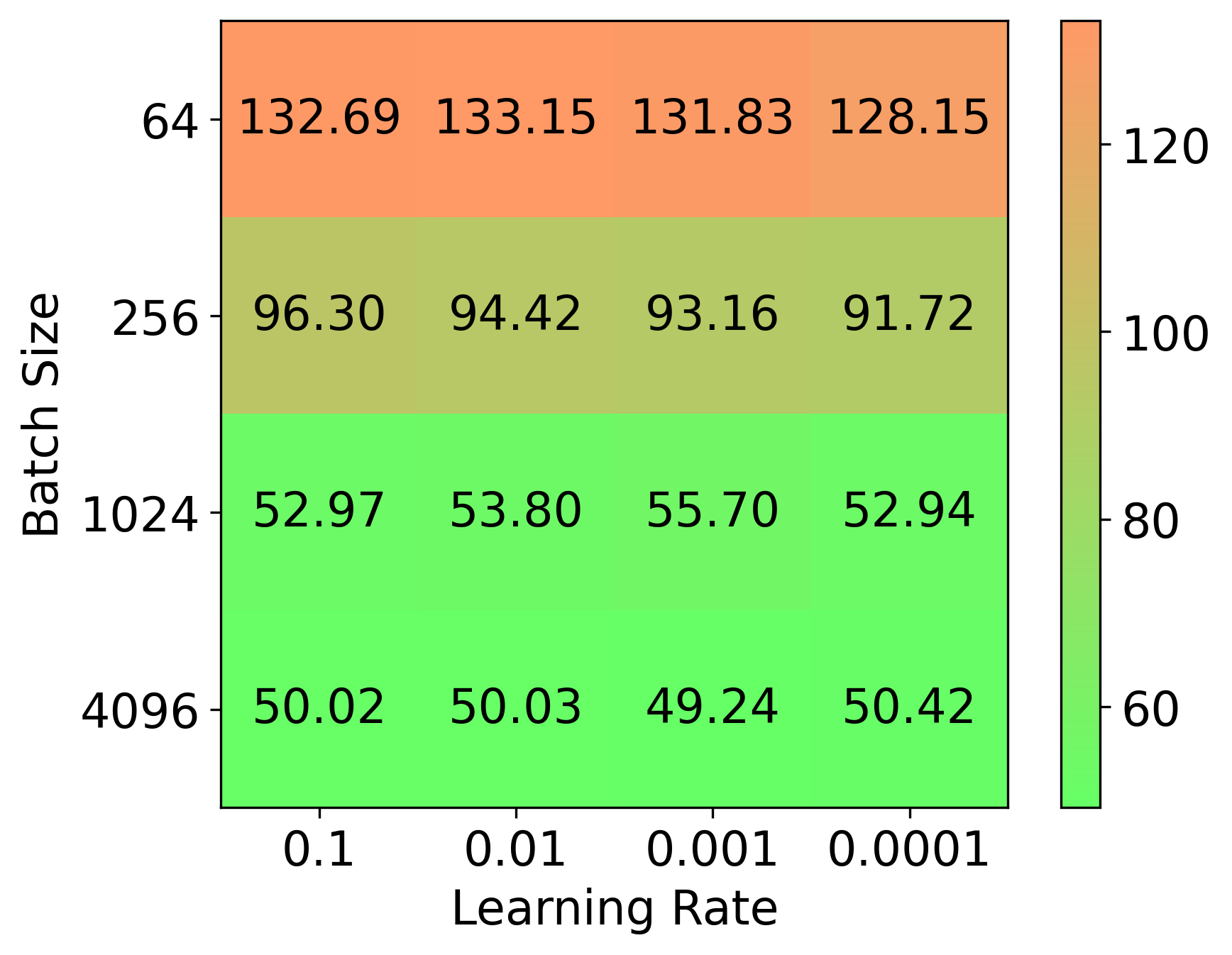}\label{fig:pam_h100_1}}
\end{minipage}
\hfill
\begin{minipage}[t]{0.32\textwidth}
  \centering
  \subfloat[PAMAP2 on A100]{\includegraphics[width=\linewidth]{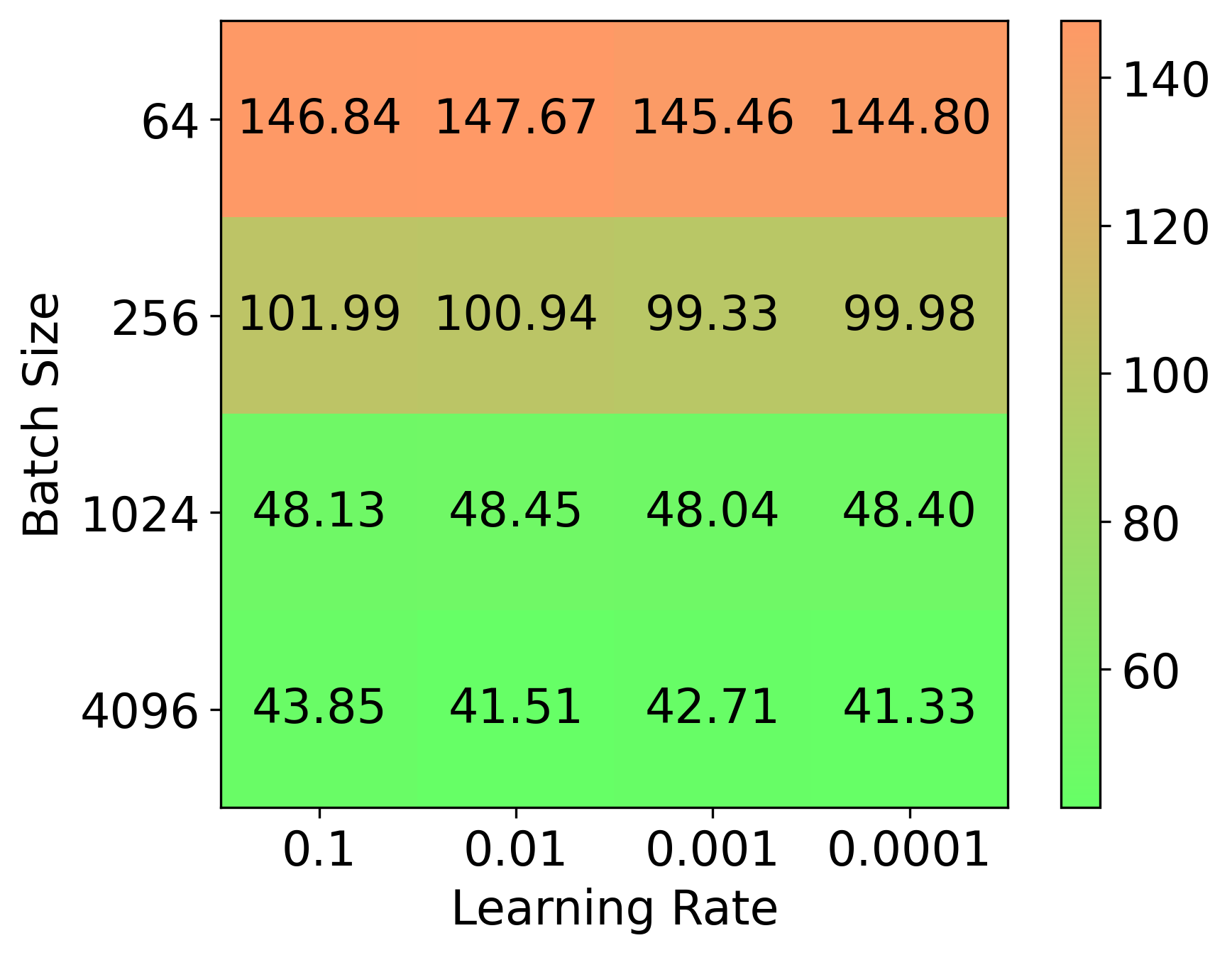}\label{fig:pam_a100_1}}
\end{minipage}
\hfill
\begin{minipage}[t]{0.32\textwidth}
  \centering
  \subfloat[PAMAP2 on RTX 3090]{\includegraphics[width=\linewidth]{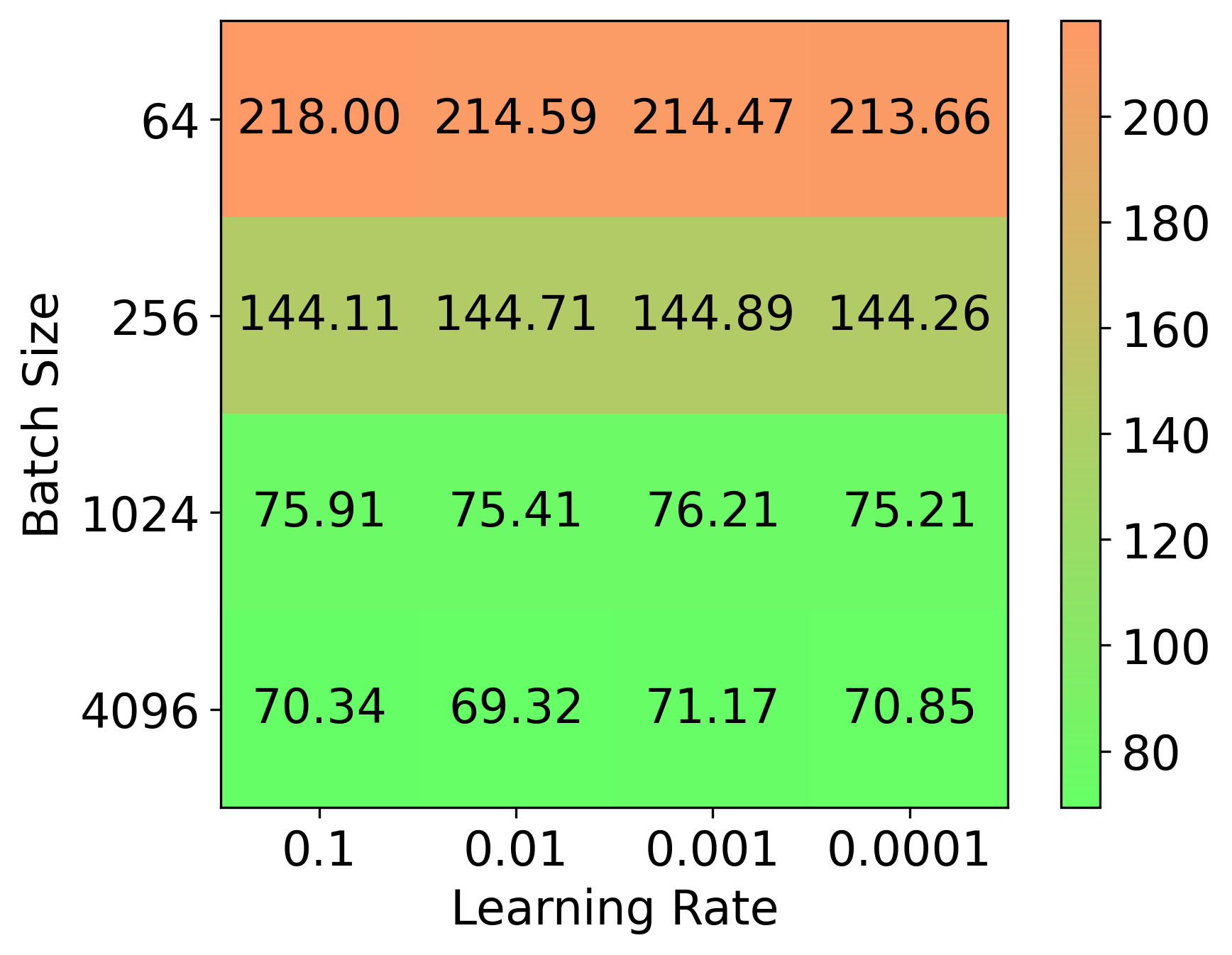}\label{fig:pam_RTX_1}}
\end{minipage}

\caption{Energy per epoch (mWh) across the different application scenarios with varying Batch Size and Learning Rate.}
\label{fig:energy_epoch}
\end{figure}

\begin{figure}[!ht]
\begin{minipage}[t]{0.48\textwidth}
  \centering
  \subfloat[CelebA on Nvidia H100]{\includegraphics[width=\linewidth]{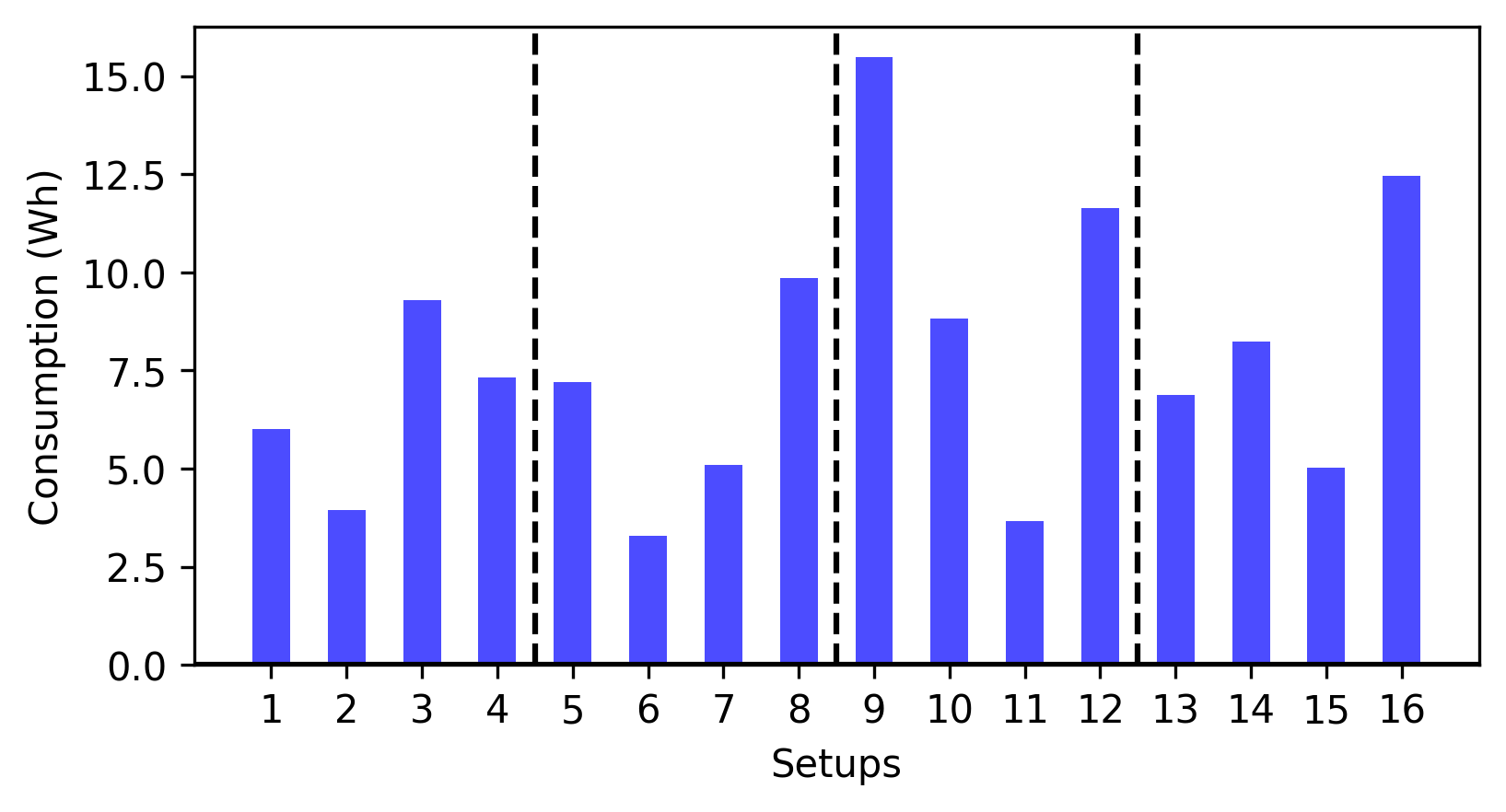}\label{fig:cel_H100}}
\end{minipage}
\hfill
\begin{minipage}[t]{0.48\textwidth}
  \centering
  \subfloat[PAMAP2 on Nvidia H100]{\includegraphics[width=\linewidth]{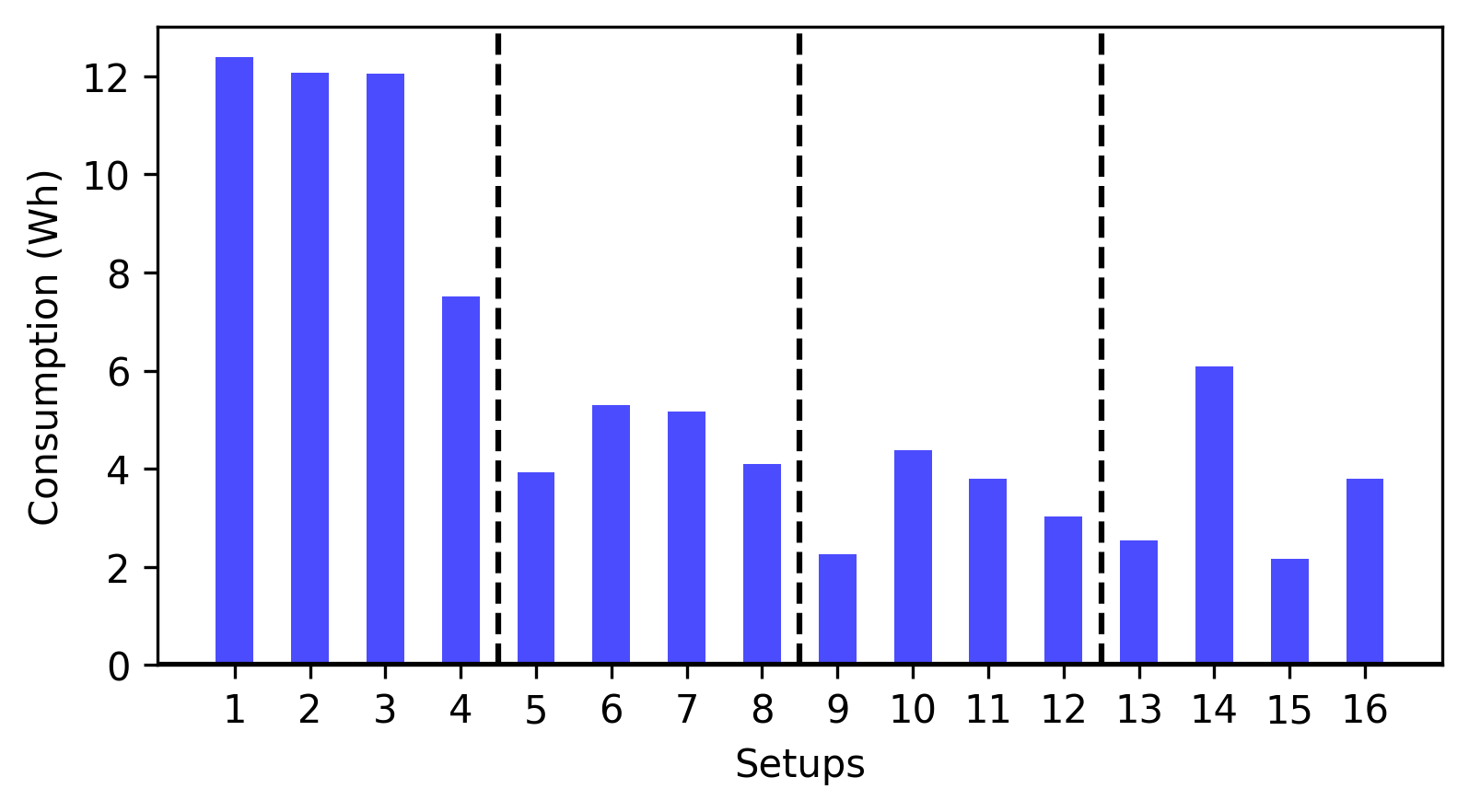}\label{fig:pam_H100}}
\end{minipage}
\vspace{5px}

\begin{minipage}[t]{0.48\textwidth}
  \centering
  \subfloat[CelebA on Nvidia A100]{\includegraphics[width=\linewidth]{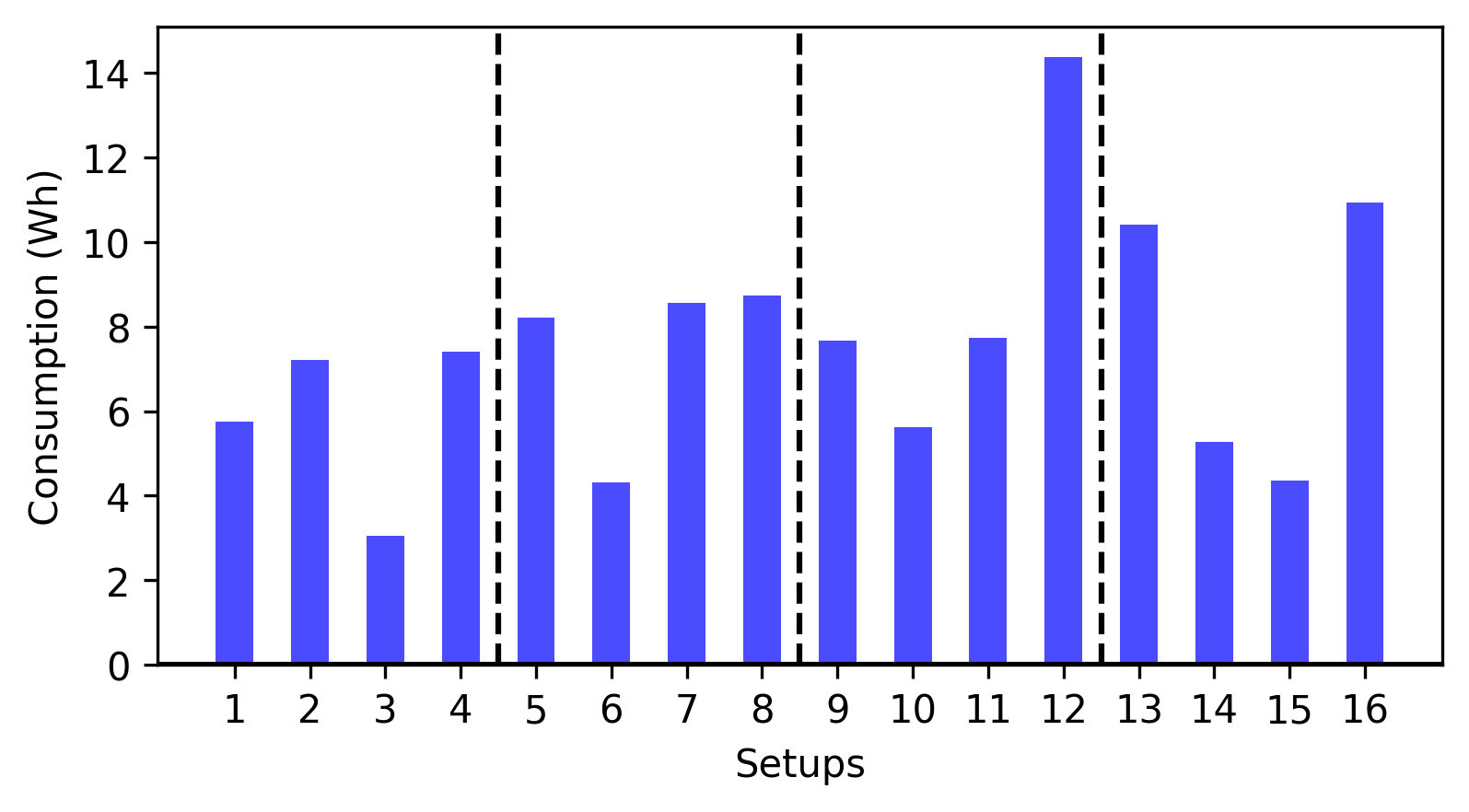}\label{fig:cel_A100}}
\end{minipage}
\hfill
\begin{minipage}[t]{0.48\textwidth}
  \centering
  \subfloat[PAMAP2 on Nvidia A100]{\includegraphics[width=\linewidth]{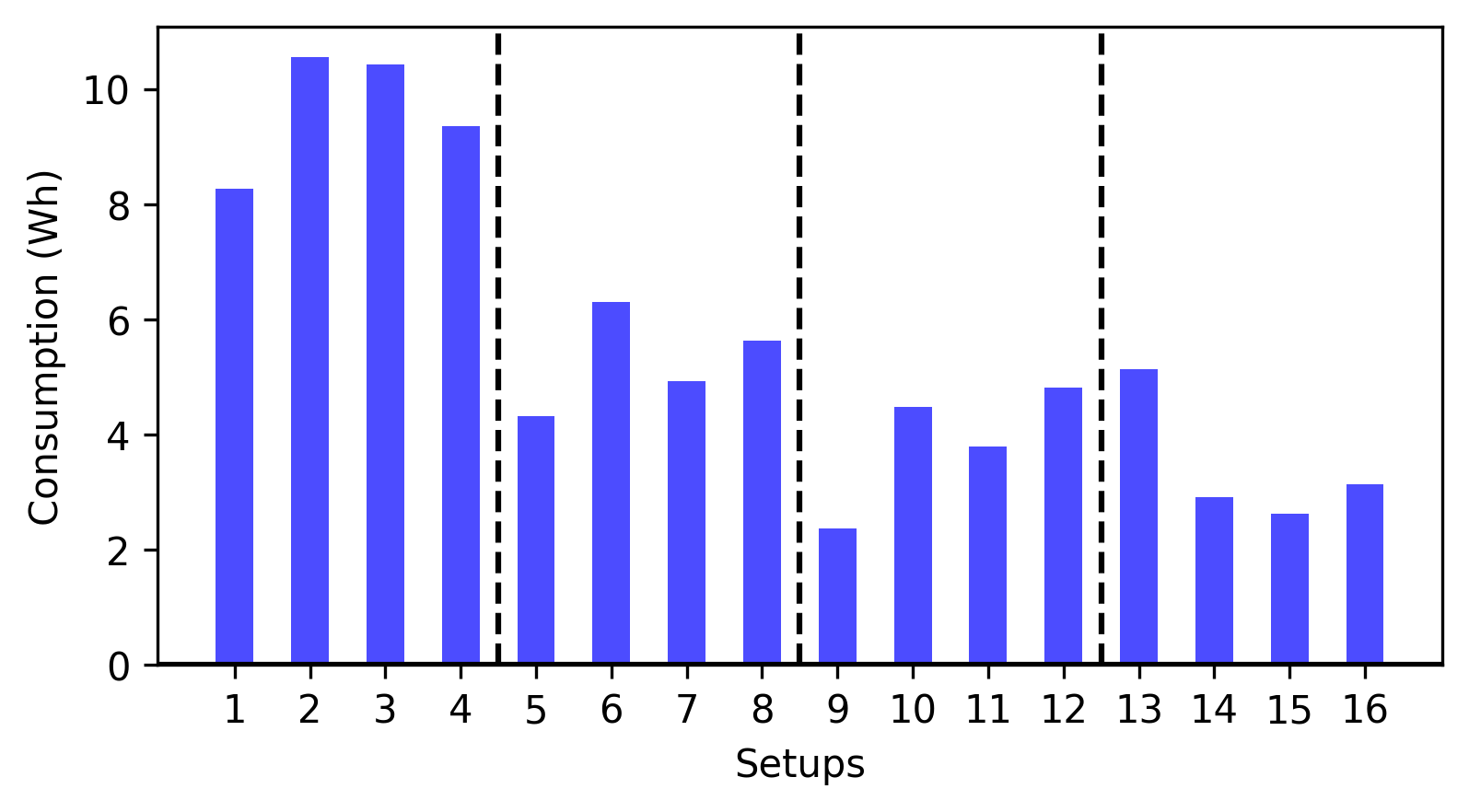}\label{fig:pam_A100}}
\end{minipage}
\vspace{5px}

\begin{minipage}[t]{0.48\textwidth}
  \centering
  \subfloat[CelebA on Nvidia RTX 3090]{\includegraphics[width=\linewidth]{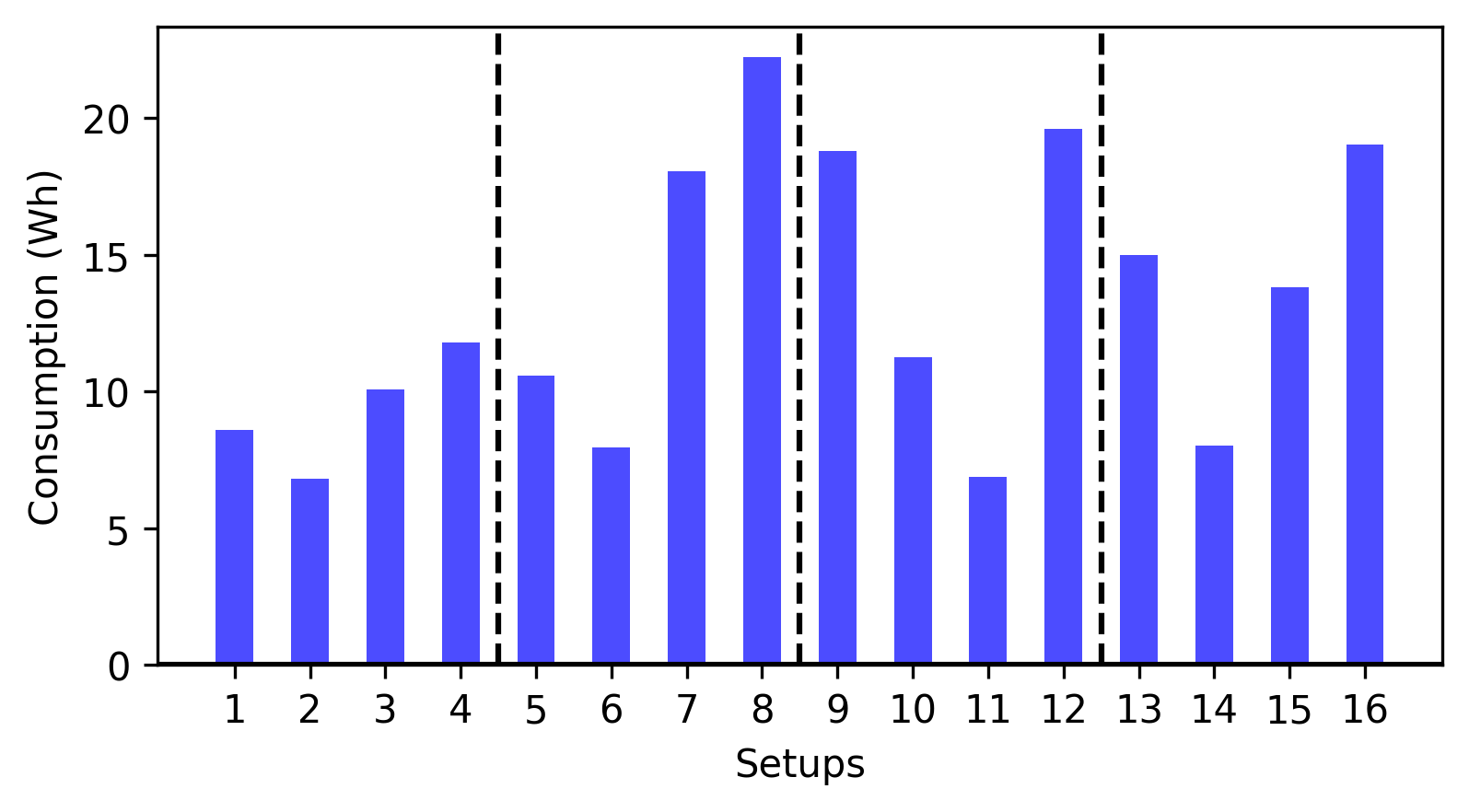}\label{fig:cel_rtx}}
\end{minipage}
\hfill
\begin{minipage}[t]{0.48\textwidth}
  \centering
  \subfloat[PAMAP2 on Nvidia RTX 3090]{\includegraphics[width=\linewidth]{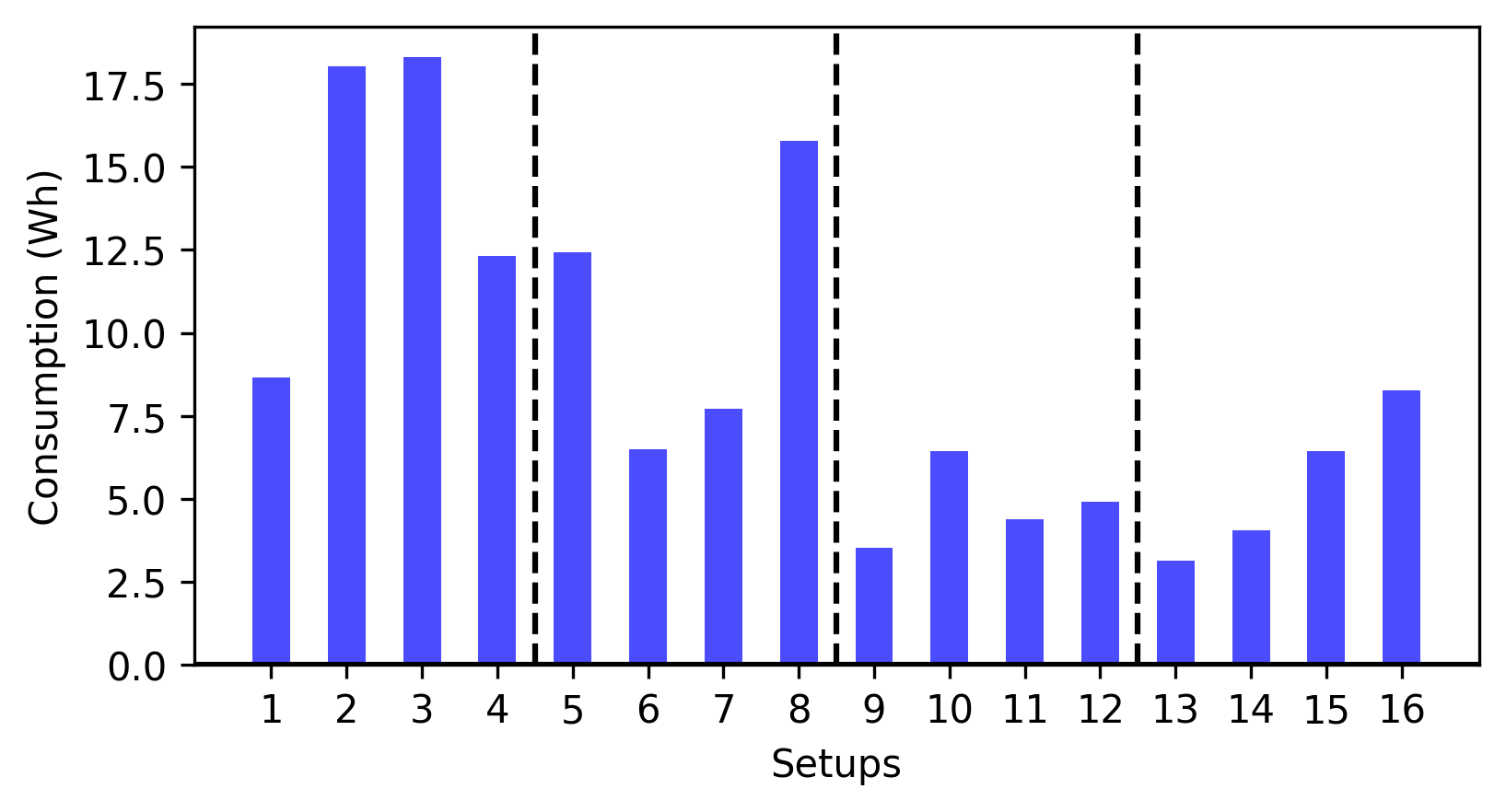}\label{fig:pam_rtx}}
\end{minipage}

\caption{Average consumption across all epochs for each application scenario and test setup.}
\label{fig:average_energy}
\end{figure}

For testing the training regime on the two application scenarios, the standard versions of their tasks are utilized by excluding the advanced setups so far.
For PAMAP2, this is the training of an encoder and a classifier in one run without pretraining, whereas, for the CelebA scenario, a single classifier is trained to predict the identities only.
Each scenario is trained with 16 different hyperparameter combinations and run two times on the three different HPC hardware setups.

The average energy that was demanded during the training is shown in \Cref{fig:energy_epoch}, calculated through the introduced equation \Cref{equation1}.
For this equation, k states the number of epochs in range 0 to T and n represents the number of power measurements within an epoch from which we calculate the average.
As a general rule, the level of energy used can be linked to the efficiency and utilization of the GPU.
Within each matrix, lower values mean less energy consumed for training one epoch, which results in improved GPU efficiency.

For each matrix shown in table \Cref{fig:energy_epoch}, the batch size choice has the most impact on the energy consumption.
Since the batch size determines the level of operations that can be executed in parallel, effects in the energy per epoch mainly originate from the utilization and the efficiency window in which the GPU is operated.
There needs to be a balance between the dataset size and the GPU capabilities in order to operate it efficiently.
Cross-comparing the matrices across the two application scenarios highlights the importance of proper hyperparameter utilization.
Due to the large size of the CelebA image dataset with each data point representing a 78×218 pixel-sized image, the smallest batch size can be considered the best selection to operate the GPU efficiently.
On the other hand, the tremendously smaller size of the PAMAP2 results in in-efficient utilization of the GPU for small batch sizes.
Therefore, larger batch sizes run more efficiently for the PAMAP2 scenario since less energy is wasted for GPU idling.

Compared to the batch size, the choice of learning rate does not have a recognizable influence in this analysis.
Apart from small deviations, which may occur due to the software-based measurement procedure, we cannot derive a trend for the energy per epoch across the four learning rate selections.
The reason for that is the independence of the training duration, namely the number of epochs until the model converges and stops the training process.
As mentioned before, the early stopping was set to start after reaching the threshold accuracy of 0.8 with patience of 40 epochs. 
Influences through the learning rate change can therefore be seen in figure \Cref{fig:average_energy}.
Across the 16 setups, grouped into the 4x4 setup alignment through the dashed lines as introduced in \Cref{tab:parameters}, the total energy consumption for the full training is visualized.
We can derive a slight trend of learning rate settings towards 0.01 and 0.001, due to the issue of learning rate 0.1 being set too large to converge properly whereas a learning rate of 0.0001 increases the number of epochs needed until convergence.
Further, the total consumption is still ruled by the batch size choice, especially in the PAMAP2 scenario with greater deviations in energy per epoch.

Overall, the range between the worst and best-performing setup is wide, with the best case usually consuming less than 20\% of the worst-case setup.
As a guideline to achieve this much of an improvement, the energy per epoch and necessary number of epochs both need to be minimized.
Comparing our two application scenarios, especially when checking the CelebA results in \Cref{fig:average_energy}, there can not be derived a general rule for thoroughly choosing the best hyperparameters.
These findings therefore provide a motivation for further research to explore more dynamic and thorough hyperparameter selection approaches.

\section{Pretraining Learning Paradigm Results }
\label{sec:pre_results}

For the pretraining scenario of PAMAP2, the autoencoder's energy consumption for training the encoder was tracked first.
On top of that, the consumption for training a classifier based on the latent features was tracked as well.
As a comparison, we trained the encoder and classifier architecture together without pretraining.
In total, we trained each model again for all 16 setups.

To compare the results, we generated equation \ref{equation2} to calculate the breakeven point on how many times the pre-trained encoder model needs to be recycled in order to compensate the pretraining energy overhead compared to training the full architecture every time.

\begin{equation}
\begin{aligned}
  &E_{Enc} +x* E_{Class} =  x * E_{Enc+Class}\\
  &x = \frac{E_{Enc}}{ ( E_{Enc+Class}-E_{Class})}\\
  &\text{with x = number of cycles till compensation}
  \end{aligned}
  \label{equation2}
\end{equation}
\vspace{5px}

\begin{figure}[!ht]
\begin{minipage}[t]{0.4\textwidth}
  \centering
  \subfloat[PAMAP2 on Nvidia H100]{\includegraphics[width=\linewidth]{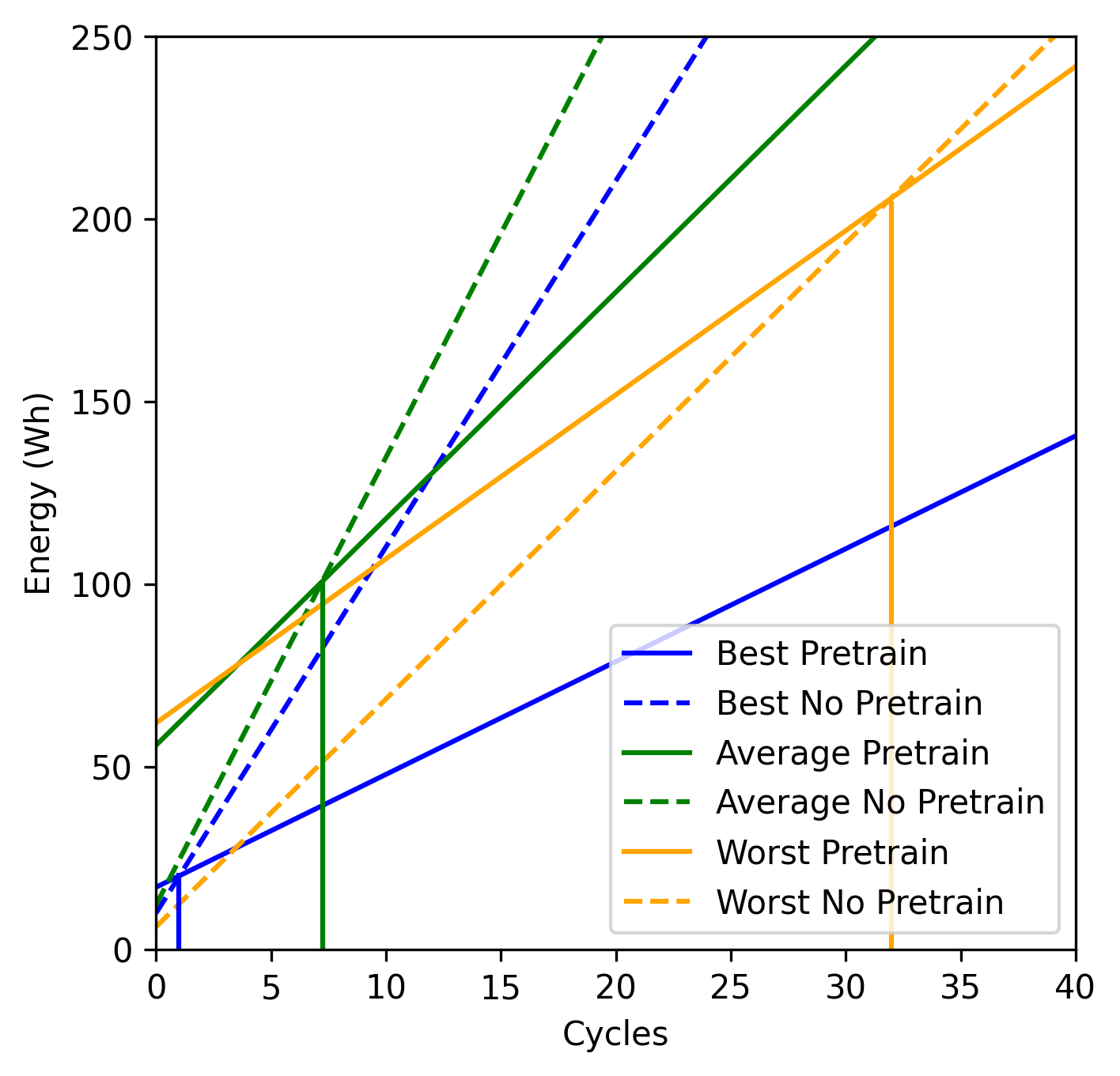}\label{fig:break_h100}}
\end{minipage}
\hfill
\begin{minipage}[t]{0.4\textwidth}
  \centering
  \subfloat[PAMAP2 on Nvidia A100]{\includegraphics[width=\linewidth]{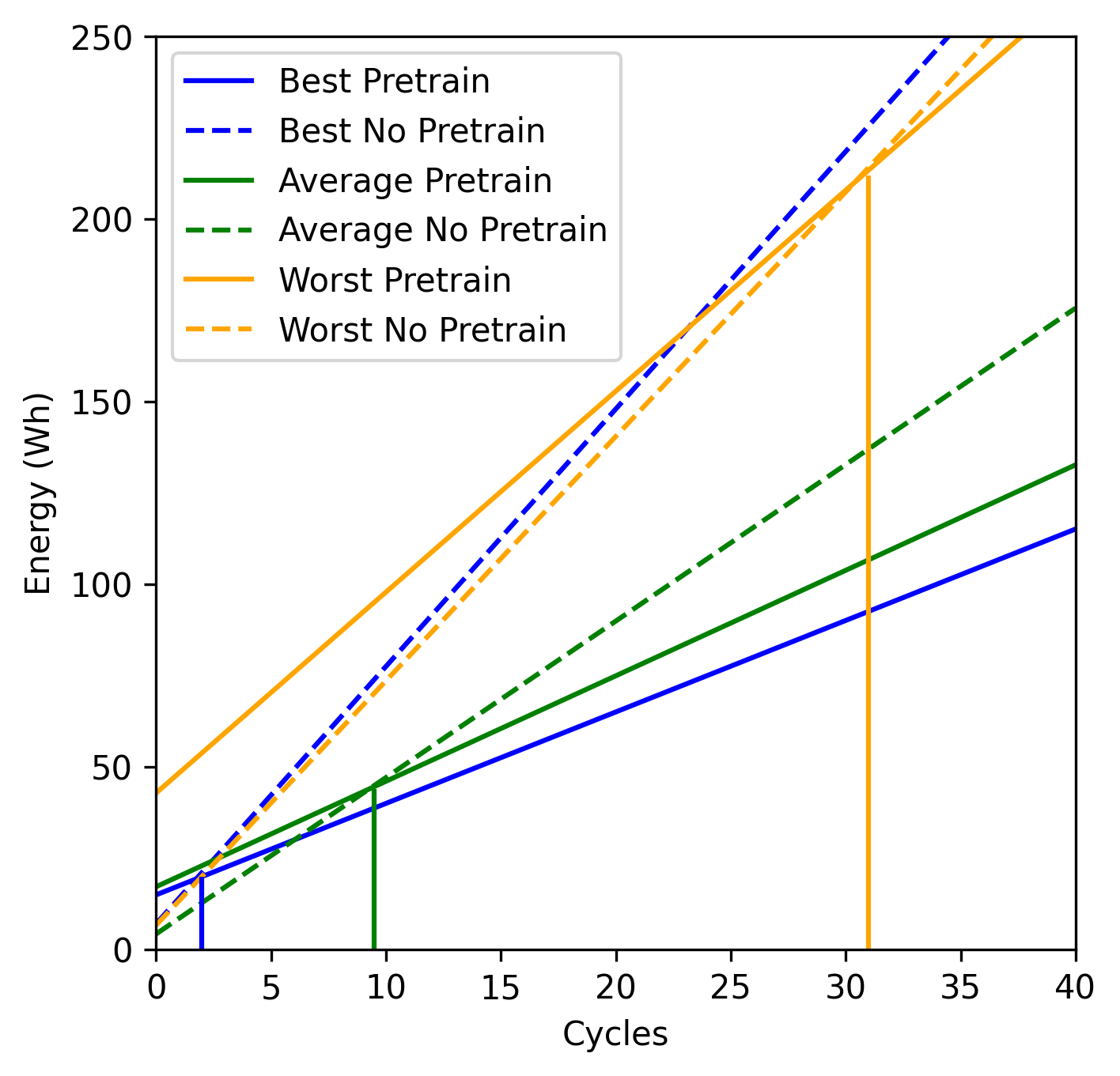}\label{fig:break_a100}}
\end{minipage}

\begin{center} 
\begin{minipage}[t]{0.4\textwidth} 
  \centering
  \subfloat[PAMAP2 on Nvidia RTX 3090]{\includegraphics[width=\linewidth]{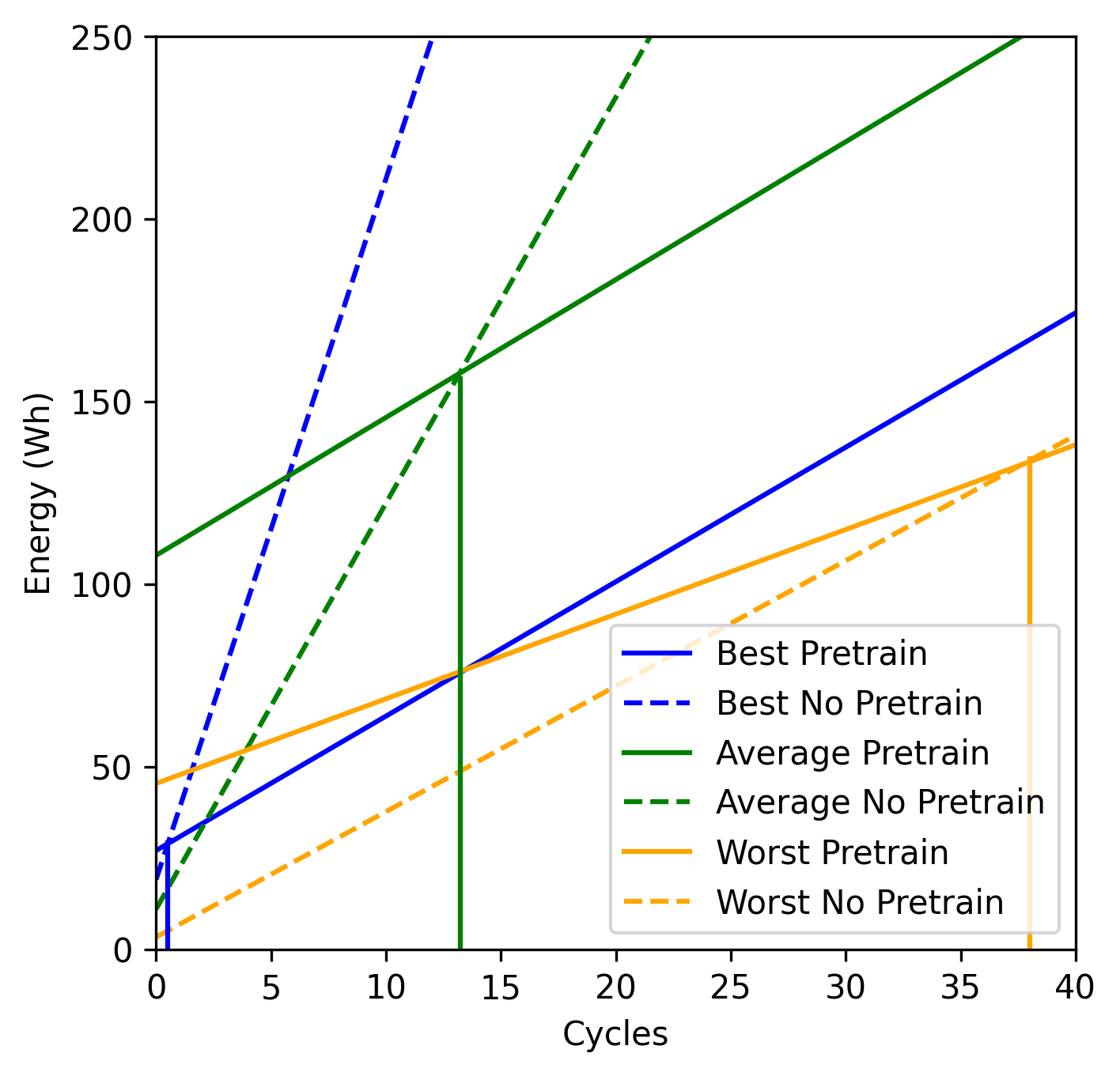}\label{fig:break_rtx}}
\end{minipage}
\end{center}
\caption{Break-even points, highlighting the amount of cycles until the pretraining energy investment is compensated.}
\label{fig:break_even}
\end{figure}

The results are shown in \Cref{fig:break_even} for the three desired hardware setups.
In this case, the results are solely based on the frozen encoder, so the optimization process does not consider the encoder anymore.
The consumption for the pretraining run is shown in solid lines whereas the basic training without encoder pretraining is a dashed line.
The crossing of two lines of the same color indicates the break-even point, which means with how many complete downstream-task training cycles, the investment in the pretraining starts to yield better energy savings in the longer perspective.
The dashed pre-trained lines commonly start at a higher energy level since the starting point represents the pretraining energy investment, whereas the slope of each line represents the addition of the recycling cycle's training energy over time.
The ideal situation is that the break-even point appears as early as possible.
From the 16 setups of hyperparameter configurations utilized, we extracted the best, worst, and average scenarios.

As we can see across the three hardware setups, the blue lines show a scenario where the pretraining already outperforms the standard training after two to three recycling steps.
On the other hand, for the worst case, visualized with orange, it takes between 30 to 40 cycles until the spent pretraining energy is compensated.
As we can derive from equation \ref{equation2}, the compensation depends on the increased energy consumption of the encoder pretraining and the ratio between training the classifier alone or with the encoder together.
On average, the energy is compensated after seven to 13 recycling steps of the pre-trained encoder.
Interestingly, in \Cref{fig:break_rtx}, the worst-case scenario with almost 40 recycling steps still consumes less total energy than the average recycling.
The reason for that is the focus on cycles instead of energy in this graphic, leading to the approval that even though you can recycle the energy with fewer cycles, the total consumption has to be taken into account as well in order to maintain the life-cycle efficiency.

Additionally, we investigated the average energy consumption between freezing and unfreezing the encoder's layer weights after its pretraining run. \cite{marcelino2018transfer}
Since the optimizer does not update the weights during the backpropagation to fine-tune the encoder, there is less energy needed to train the full model.
As shown in  \Cref{freezeUnfreeze}, average energy saving between 23\% to 45\% can be achieved across all 16 setup combinations.
Since proper pretraining already ensures the required encoders' behavior, the freezing of weights can be a helpful action to reduce energy consumption, and further energy investment for fine-tuning can be saved.

\begin{table}[ht]
    \caption{Average Energy consumption (Wh) for utilizing the pre-trained encoder in frozen or unfrozen mode.}
    \label{freezeUnfreeze}
    
    \centering
    \renewcommand{\arraystretch}{1.25} 
    \setlength{\tabcolsep}{5pt} 
    \begin{tabular}{c|ccc}
        \multicolumn{1}{c}{} & \multicolumn{3}{c}{\textbf{Hardware}} \\
        \cline{2-4}
        \textbf{Encoder} & \text{H100} & \text{A100} & \text{RTX3090}\\
        \hline  
        \text{Frozen} & 3.89 Wh & 4.24 Wh & 4.97 Wh \\
        \text{Unfrozen} & 6.01 Wh & 5.55 Wh & 9.13 Wh \\
        \hline
        \text{Savings} & 35.2\% & 23.6\% & 45.5\% \\
    \end{tabular}
    
\end{table}

\section{Multitask Learning Paradigm Results }
\label{sec:mutli_results}

For the multitask scenario trained on CelebA, we utilize the ResNet18 architecture pre-trained on ImageNet as the encoder.
Due to ImageNets size and required training time, we decided to recycle an already existing, pre-trained model.
As a side note, the best option to save energy in machine learning is the usage of already existing, pre-trained models from external, shared sources to compensate their energy investment further.

For our experiment, the nature of the CelebA dataset with multiple properties labeled per image, we focus on training a multitask classifier and measuring the energy consumption including the pre-trained encoder.
More specifically, we train a classifier to classify 40 facial attributes and the 40 most represented person identities of a CelebA image.




The experiment was conducted through the same procedure as the previous pretraining scenario, training the three models (attribute, identity, and both together) with the 16 different setups.
We calculated the average energy per epoch across the three scenarios as shown in \Cref{celeb_type_energy}.
Since the average energy per epoch is almost identical across training attributes, identity, or both, with different operating levels across the three hardware configurations, we further calculated the average length of the training in number of epochs.
As we can see, for this experiment the number of epochs rules the overall energy consumption whereas the efficiency from energy per epoch is negligible.
Since the loss functions are connected during the training in the multitask scenario, the classifiers of attribute and identity are able to cross-regulate each other, which results in a quicker convergence.
Instead of recycling already present knowledge as for the pretraining, this experiment benefits from the active exchange of information through the loss function.
The result manifests our previous findings for minimizing either one or at best both, the energy per epoch and the overall number of epochs, to properly decrease the energy consumption for the training.

\begin{table}[ht]
\caption{Average Energy per Epoch (mWh) / Average number of Epochs }
    \label{celeb_type_energy}
    \vspace{5px}
    \centering
    \renewcommand{\arraystretch}{1.25} 
    \setlength{\tabcolsep}{5pt} 
    \begin{tabular}{c|ccc}
        \multicolumn{1}{c}{} & \multicolumn{3}{c}{\textbf{Hardware}} \\
        \cline{2-4}
        \textbf{Type} & \text{H100} & \text{A100} & \text{RTX3090}\\
        \hline  
        \text{Attribute} & 47.93 mWh / 358 & 46.82 mWh / 380 & 81.41 mWh / 382 \\
        \text{Identity}  & 47.26 mWh / 170 & 45.79 mWh / 163 & 81.42 mWh / 159 \\
        \text{Both}      & 48.34 mWh / 85  & 45.48 mWh / 86  & 81.19 mWh / 84 \\
    \end{tabular}
    
\end{table}

\section{Correlation with High-End Consumer Hardware}
\label{sec:prosumer}

We transferred our experiment to high-end consumer hardware to compare our findings independently of the utilized HPC cluster.
Therefore we ran our experiment scenarios for the CelebA and the PAMAP2 dataset again on a workstation powered by an Nvidia RTX 6000 Ada with 48GB VRAM.
Further, to test the independence of CUDA-based systems, we set up a 16-inch Apple Macbook Pro with M1 Max chip with 32-core GPU and 64GB unified memory as a second device. 
The Nvidia GPU workstation runs on CUDA 12.3 and the Apple device runs on metal performance shader (MPS) for neural network training acceleration.
We implemented a custom tracker based on the \textit{powermetrics} tool of Mac OS to generate the same energy monitoring and data structure as collecting data through Carbontracker.

For the training regime, we could reproduce the influences of batch size and the learning rate influence on the energy per epoch efficiency metric.
However, the findings were somewhat less distinctive and showed higher variations.
We attribute this to the missing isolation of the jobs, which are now running on commercial operating systems.
Additionally, for the powermetrics-based tracking, it remains unclear how precise the measurements can be compared to the carbontracker measurements.
Even though the local hardware operated on different power levels, running the two learning paradigm tests confirmed our findings that pretraining and multitask learning can be utilized to improve life-cycle-oriented efficiency.
For pretraining, the average breakeven points are calculated for 9 cycles on the CUDA setup and 12 cycles on MPS.
Finally, for the multitask learning, the information sharing through the loss function by training attribute and identity at the same time could reduce the energy consumption by around 40\%.

\section{Conclusion}
\label{sec:conlusion}
In conclusion, we investigated the energy consumption of two popular ML scenarios on three different HPC and 2 pro-consumer workstation configurations with variations in batch size and learning rates.
For the 16 selected setups, we first evaluated their influence on the efficiency through energy per epoch and further, the total energy per training cycle until the early stopping monitor criteria were met.
Our findings showcased up to 80\% decrease in energy consumption if hyperparameters are set properly.
For the learning paradigm tests, the calculated break-even point is an indicator if energy investment for pretraining is beneficial.
Proper selection of hyperparameters may help to compensate for the energy overhead already after a few recycling steps.
A long recycling process can be necessary if too much energy is invested into the pretraining or if the savings are too weak for each recycling step.
Lastly, for the multitask training, we not only showed that training two classifiers in one training process is less energy-consuming, we additionally showed that fewer epochs are necessary in total due to the mutual training support of the classifiers.

As a limitation, we observed that hyperparameters influence the pretraining and downstream task cycles differently, this is mostly determined by the number of epochs needed to reach the accuracy target. 
Therefore, it is not possible to derive best practice recommendations without trying them out.
Moreover, even the dataset influences the energy consumption depending on its size and the chosen number of samples for pretraining and thorough training.
Nevertheless, as the goal of training DNNs is to have a model that reaches good performance, optimizing for power consumption alone is not enough. 
Future research in this direction should balance the model prediction performance and the energy efficiency based on the baseline findings from our work.

\section*{Acknowledgement}
This work is supported by the European Union’s Horizon Europe research and innovation program (HORIZON-CL4-2021-HUMAN-01) through the "SustainML" project (grant agreement No 101070408).

%
%
%
\bibliographystyle{splncs04}
\bibliography{references}
%




\end{document}